# Real-time Texture Error Detection

**Assoc. Prof. Ph. D. Eng. Dan Laurenţiu Lacrămă**
**„Tibiscus" University of Timişoara, România**
**Assoc. Prof. Ph. D. Eng. Florin Alexa**
**Politehnica University of Timişoara, România**
**Lect. Ph. D. Eng. Adriana Baltă**
**„Tibiscus" University of Timişoara, România**

**Abstract:** This paper advocates an improved solution for the real-time error detection of texture errors that occurs in the production process in textile industry.
The research is focused on the mono-color products with 3D texture model (Jacquard fabrics). This is a more difficult task than, for example, 2D multicolor textures.

## 1. Introduction

Textile industry need, automatic real-time quality control in order to avoid end products mistakes in a quick end efficient manner. Manual control is inefficient, time consuming and implies most times heavy materials losses.

On another hand automatic quality control is much more efficient, because it is real-time, implies almost no material losses and no workers but its effectiveness is directly dependent on some technical matters. Most important of them are:

- Optical sensors quality;
- Controlled illumination;
- Software ability to quickly detect fabric's errors.

The block scheme of an automatic quality controlled production process is depicted in Figure 1.1.

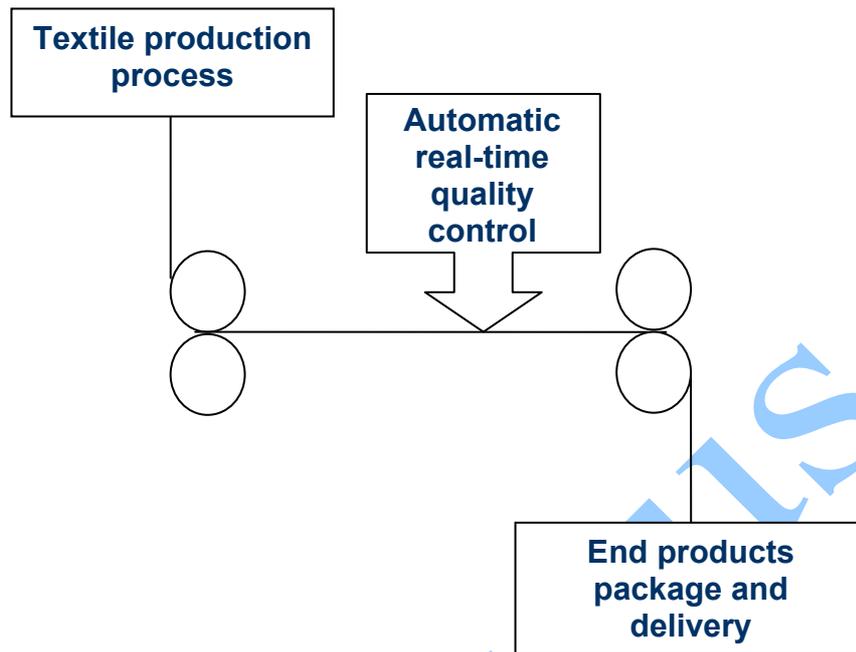

**Figure 1.1.** General structure of an automatic quality control system

Automatic error detection is performed by a specially design software able to compare in real time the fabric's image with the desired pattern of the textile material. It looks simple, but it is not because a lot of problems must be solved and the procedure must be robust accordingly to a lot of technological parameters variations inside normal limits.

Therefore, images' acquisition is a highly important step for the automatic quality control because it provides the input data for the whole process. The acquisition is performed by an optical sensor which is almost always a video camera (with one line or a matrix of CCD) able to provide directly good quality digital images. It is important for these images to be accurate and noiseless; hence the preprocessing procedures needed on them are quick, simple and no data loss or error giving.

Local illumination is also directly linked with the quality image acquisition because it is straight forward to demonstrate that its variations can heavily affect the patterns visibility in the taken fabrics photos and videos.

Consequently the natural sources of light which are non-constant must not be employed and even more their influence should be carefully eliminated. Thus, the use of a strictly controlled illumination provided exclusively by one or more artificial light sources is the single reasonable alternative.

The software designed to perform the image processing and the errors detection has some sub-modules able to perform specific tasks such as:

a.   Image preprocessing;
b.   Image transform;
c.   Feature extraction;
d.   Matching with the model;
e.   Error detection.

**a. Image preprocessing** -even if we use good quality cameras together with an adequate artificial illumination, image preprocessing cannot be avoided.

Anyway the low level processes employed are not so much time consuming and the advantages in latter steps are really important. The preprocessing procedure involves:

- Brightness/Contrast corrections;
- Noise reduction (technological process provides a lot of electromagnetically noise which influence on raw images is impossible to avoid);
- Contours detections;
- Line thinning.

**b. Image transform** - in the bibliography we found lot of proposed solutions for the image transform. Most frequent are:

- Cosine Transform;
- Fast Fourier Transform;
- Wavelets.

**c. Feature extraction** – after the previously discussed transformation the resulting images can be described by a group of parameters from which we have to select the most relevant for our purpose. For example if we use a Cosine Transform it is important to select as many coefficients as needed to be able to do a reliable matching with the ideal model pattern. This means more than 99% of the errors to be detected and no "false alarms" (correct patterns declared as erroneous by the automatic quality control). The last condition is also very important, because a "false alarm" means stopping production for no reason and thus lot of time and energy wasted.

**d. Matching with the model** - can be done using all the methods provided by the pattern recognition techniques. If the extracted features are statistical, as it is usually the case, statistic matching is performed. If the features are syntactic, than the linguistic theory is employ to detect similarity or non similarity. A third solution is to combine both and use some complex procedure of matching, for example a neural network.

**e. Error detection** – after measuring the degree of matching between the current fabric picture and the ideal texture pattern a decision is taken whether the differences are inside the normal limits or an error occurred. It the second situation, production process must be quickly stopped and corrections need to be done in order to eliminate the error causes.

Our research was focused on a particular problem which raises some extra difficulties from the usual cases: Monochromatic 3D texture fabrics, also called Jacquard textile from the name of the inventor of the machine that produce them. An example of such a textile fabric is given in Figure 1.2.

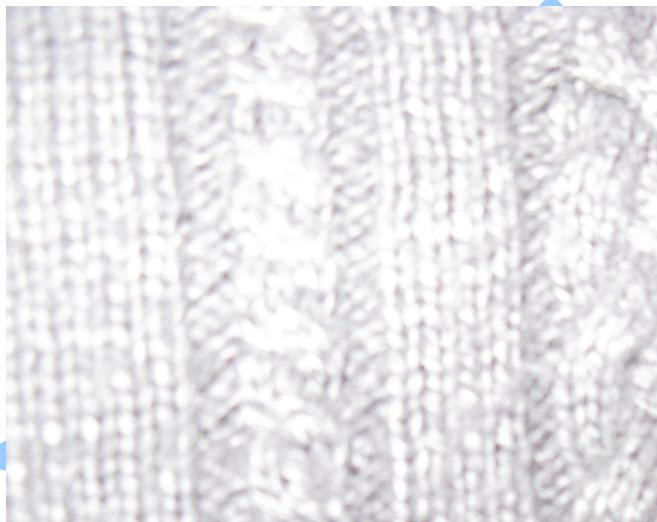

**Figure 1.2.** Monochromatic 3D fabric

## 2. Proposed solution

Our proposed solution for this problem is sketched in Figure 2.1. The differences from the standard solution are:

- two video camera instead of one in order to take account of the 3D texture;
- the natural illumination is completely eliminated, thus an artificial controlled illumination is employed;
- the Hough transform is used for extracting the recognition features;

- a Multilayer perceptron is takes the final decision over the error detection.

The general schema is provided in Figure 2.1. It is focused mainly on the expert software because it is this paper main topic.

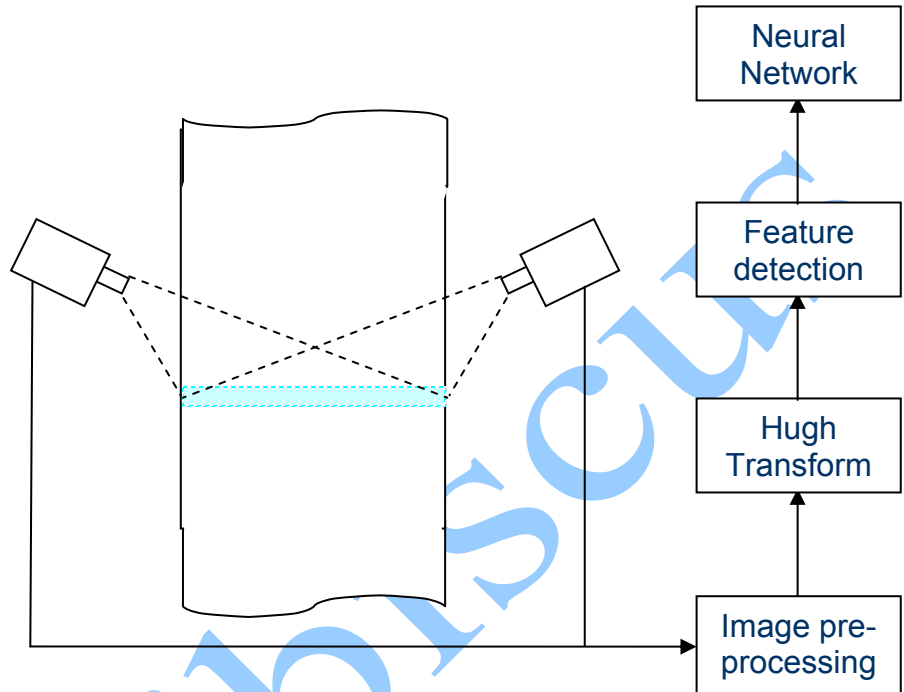

**Figure 2.1.** The proposed solution for the automatic quality control

The images taken by the two video cameras undergo the same preprocessing procedures:

- Gauss filtering for noise reduction;
- Laplace filtering for contours detection;
- Line thinning in order to simplify the recognition.

A comparative research was made between two procedures with and without noise reduction. Final results proved that the risk of false alarm is significantly bigger if no noise filtering is done.

After this stage, the resulting images are two angle views of the fabric skeleton containing only one pixel thick lines.

All the above mentioned transforms can be employed for the multi-colored 2D texture products quality control, but for monochromatic fabrics the best results were obtained with the Hough transform. It is much more linked with lines and directions coding. In the end of this transform the images are replaced by Cartesian graphical representations of line direction densities.

Theoretically we could directly employ the two images Hough transform as inputs for the Neural Network, but that would have brought the necessity of a very big neural architecture with low speed. During experiments we found out that using only statistical data leads to even better results in less time. Statistical data has the advantages of not being very sensible to small variations that are normal in the industrial production area.

Therefore we employed:
- The average value
- The minimum and maximum values together with their position on horizontal axe.
- The standard deviation

Experiments were carried out with both Multilayer Perceptron and Radial Bassis Neural networks. NN implementation was done using the NeuroSolutions4 software developed by NeuroDimension Inc.

Multilayer perceptrons with one hidden layer were the most efficient so far. A MLP standard structure is sketched in Figure 2.2.

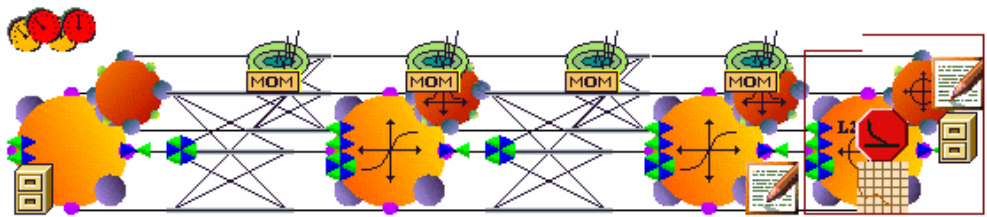

**Figure 2.2.** Multilayer perceptron

## Conclusions

The proposed solution proved to be quick quite effective and rather inexpensive but we intend to continue our research in order to obtain similar good results in even shorter time. This is very important because each second of delay between error occurrence and the process stop means wasted material and therefore more effort must be done.